# HYPERGRAPH BASED SEMI-SUPERVISED LEARNING ALGORITHMS APPLIED TO SPEECH RECOGNITION PROBLEM: A NOVEL APPROACH

Hoang Trang[1], Tran Hoang Loc[1]

[1] *Ho Chi Minh City University of Technology-VNU HCM, Ho Chi Minh City, Vietnam*

Corresponding author*: tran0398@umn.edu*

## ABSTRACT

Most network-based speech recognition methods are based on the assumption that the labels of two adjacent speech samples in the network are likely to be the same. However, assuming the pairwise relationship between speech samples is not complete. The information a group of speech samples that show very similar patterns and tend to have similar labels is missed. The natural way overcoming the information loss of the above assumption is to represent the feature data of speech samples as the hypergraph. Thus, in this paper, the three un-normalized, random walk, and symmetric normalized hypergraph Laplacian based semi-supervised learning methods applied to hypergraph constructed from the feature data of speech samples in order to predict the labels of speech samples are introduced. Experiment results show that the sensitivity performance measures of these three hypergraph Laplacian based semi-supervised learning methods are greater than the sensitivity performance measures of graph based semi-supervised learning methods (i.e. the current state of the art network-based method for classification problems) and the Hidden Markov Model method (the current state of the art method applied to speech recognition problem) applied to network created from the feature data of speech samples.

## 1. INTRODUCTION

Speech recognition is the important problem in pattern recognition research field. In this paper, we will present the hyper-graph based semi-supervised learning methods, derive their detailed regularization framework, and apply these methods to automatic speech recognition problem. To the best of our knowledge, this work has not been investigated. Researchers have worked in automatic speech recognition for almost six decades. The earliest attempts were made in the 1950's. In the 1980's, speech recognition research was characterized by a shift in technology from template-based approaches to statistical modeling methods, especially Hidden Markov Models (HMM). Hidden Markov Models (HMM) have been the core of most speech recognition systems for over a decade and is considered the current state of the art method for automatic speech recognition system [1]. Second, to classify the speech samples, a graph (i.e. kernel) which is the natural model of relationship between speech samples can also be employed. In this model, the nodes represent speech samples. The edges represent for the possible interactions between nodes. Then, machine learning methods such as Support Vector Machine (SVM) [2], Artificial Neural Networks [3], nearest-neighbor classifiers [4], or graph based semi-supervised learning methods [5] can be applied to this graph to classify the speech samples. The nearest-neighbor classifiers method labels the speech sample with the label that occurs frequently in the speech sample's adjacent nodes in the network. Hence neighbor counting method does not utilize the full topology of the network. However, the Artificial Neural Networks, SVM, and graph based semi-supervised learning methods utilize the full topology of the network. Moreover, the Artificial Neural Networks and SVM are supervised learning methods.

While nearest-neighbor classifiers method, the Artificial Neural Networks, and the graph based semi-supervised learning methods are all based on the assumption that the labels of two adjacent speech samples in graph are likely to be the same, SVM does not rely on this assumption. Graphs used in nearest-neighbor classifiers method, Artificial Neural Networks, and the graph based semi-supervised learning method are very sparse. However, the graph (i.e. kernel) used in SVM is fully-connected.

The un-normalized, symmetric normalized, and random walk graph Laplacian based semi-supervised learning methods are developed based on the assumption that the labels of two adjacent speech samples in the network are likely to be the same [5,6]. However, assuming the pairwise relationship between speech samples is not complete, the information a group of speech samples that show very similar patterns and tend to have similar labels is missed [7,8,9]. The natural way overcoming the information loss of the above assumption is to represent the data representing the set of speech samples as the hypergraph [7,8,9]. A hypergraph is a graph in which an edge (i.e. a hyper-edge) can connect more than two vertices. In [7,8,9], the symmetric normalized hypergraph Laplacian based semi-supervised learning method have been developed and successfully applied to text categorization, letter recognition, and protein function prediction applications. To the best of our knowledge, the hypergraph Laplacian based semi-supervised learning methods have not yet been applied to speech recognition problem. In this paper, we will develop the symmetric normalized, random walk, and un-normalized hypergraph Laplacian based semi-supervised learning methods and apply these three methods to the hypergraph constructed from data representing the set of speech samples available from the IC Design lab at Faculty of Electricals-Electronics Engineering, University of Technology, Ho Chi Minh City. In the other words, the hypergraph is constructed by applying k-mean clustering method to this data.

We will organize the paper as follows: Section 2 will introduce the definition hypergraph Laplacian and their properties. Section 3 will introduce the un-normalized, random walk, and symmetric normalized hypergraph Laplacian based semi-supervised learning algorithms in detail. Section 4 will show how to derive the closed form solutions of normalized and un-normalized hypergraph Laplacian based semi-supervised learning algorithm from regularization framework. In section 5, we will apply the HMM (i.e. the current state of art method applied to speech recognition problem) and graph based semi-supervised learning algorithms (i.e. the current state of art network based method applied to general classification problems) to the network created from data available from the IC Design lab at Faculty of Electricals-Electronics Engineering, University of Technology, Ho Chi Minh City and compare its sensitivity performance measure to the three hypergraph Laplacian based semi-supervised learning algorithms' sensitivity performance measures. Section 6 will conclude this paper and the future direction of researches of this speech recognition problem utilizing discrete operator of graph will be discussed.

## 2. HYPERGRAPH DEFINITIONS

Given a hypergraph $G=(V,E)$, where $V$ is the set of vertices and $E$ is the set of hyper-edges. Each hyper-edge $e \in E$ is the subset of V. Please note that the cardinality of e is greater than or equal two. In the other words, $|e| \geq 2$, for every $e \in E$. Let $w(e)$ be the weight of the hyper-edge e. Then $W$ will be the $R^{|E|*|E|}$ diagonal matrix containing the weights of all hyper-edges in its diagonal entries.

### 2.1 Definition of incidence matrix H of G

The incidence matrix $H$ of $G$ is a $R^{|V|*|E|}$ matrix that can be defined as follows

$$h(v,e) = \begin{cases} 1 \text{ if vertex } v \text{ belongs to hyperedge } e \\ 0 \text{ otherwise} \end{cases}$$

From the above definition, we can define the degree of vertex v and the degree of hyper-edge e as follows

$$d(v) = \sum_{e \in E} w(e) * h(v,e)$$
$$d(e) = \sum_{v \in V} h(v,e)$$

Let $D_v$ and $D_e$ be two diagonal matrices containing the degrees of vertices and the degrees of hyper-edges in their diagonal entries respectively. Please note that $D_v$ is the $R^{|v|*|v|}$ matrix and $D_e$ is the $R^{|e|*|e|}$ matrix.

### 2.2 Definition of the un-normalized hypergraph Laplacian

The un-normalized hypergraph Laplacian is defined as follows

$$L = D_v - HWD_e^{-1}H^T$$

### 2.3 Properties of $L$

1. For every vector $f \in R^{|V|}$, we have

$$f^T L f = \frac{1}{2} \sum_{e \in E} \sum_{\{u,v\} \subseteq E} \frac{w(e)}{d(e)} (f(u) - f(v))^2$$

2. $L$ is symmetric and positive-definite
3. The smallest eigenvalue of $L$ is 0, the corresponding eigenvector is the constant one vector 1
4. $L$ has $|V|$ non-negative, real-valued eigenvalues $0 \leq \lambda_1 \leq \lambda_2 \leq \cdots \leq \lambda_{|V|}$

Proof:

1. We know that

$$\frac{1}{2} \sum_{e \in E} \sum_{\{u,v\} \subseteq E} \frac{w(e)}{d(e)} (f(u) - f(v))^2$$
$$= \frac{1}{2} \sum_{e \in E} \sum_{\{u,v\} \subseteq E} \frac{w(e)}{d(e)} (f(u)^2 + f(v)^2 - 2f(u)f(v))$$
$$= \sum_{e \in E} \sum_{u,v \in V} \frac{w(e)}{d(e)} \left( f(u)^2 - f(u)f(v) \right) h(u,e)h(v,e)$$
$$= \sum_{e \in E} \sum_{u \in V} w(e) f(u)^2 h(u,e) \sum_{v \in V} \frac{h(v,e)}{d(e)} - \sum_{e \in E} \sum_{u,v \in V} \frac{w(e)}{d(e)} f(u)f(v)h(u,e)h(v,e)$$
$$= \sum_{e \in E} \sum_{u \in V} w(e) f(u)^2 h(u,e) - \sum_{e \in E} \sum_{u,v \in V} \frac{w(e)}{d(e)} f(u)f(v)h(u,e)h(v,e)$$
$$= \sum_{u \in V} f(u)^2 \sum_{e \in E} w(e) h(u,e) - \sum_{e \in E} \sum_{u,v \in V} \frac{w(e)}{d(e)} f(u)f(v)h(u,e)h(v,e)$$
$$= \sum_{u \in V} f(u)^2 d(u) - \sum_{e \in E} \sum_{u,v \in V} \frac{w(e)}{d(e)} f(u)f(v)h(u,e)h(v,e)$$
$$= f^T D_v f - f^T HWD_e^{-1} H^T f$$
$$= f^T (D_v - HWD_e^{-1} H^T) f$$
$$= f^T L f$$

2. $L$ is symmetric follows directly from its own definition.

   Since for every vector $f \in R^{|V|}$, $f^T L f = \frac{1}{2}\sum_{e \in E}\sum_{\{u,v\} \subseteq E} \frac{w(e)}{d(e)}(f(u) - f(v))^2 \geq 0$. We conclude that $L$ is positive-definite.

3. The fact that the smallest eigenvalue of $L$ is 0 is obvious.

   Next, we need to prove that its corresponding eigenvector is the constant one vector 1.
   Let $d_v \in R^{|V|}$ be the vector containing the degrees of vertices of hypergraph G, $d_e \in R^{|E|}$ be the vector containing the degrees of hyper-edges of hypergraph G, $w \in R^{|E|}$ be the vector containing the weights of hyper-edges of G, $1 \in R^{|V|}$ be vector of all ones, and $one \in R^{|E|}$ be the vector of all ones. Hence we have
   $L1 = (D_v - HWD_e^{-1}H^T)1 = d_v - HWD_e^{-1}d_e = d_v - HW one = d_v - Hw = d_v - d_v = 0$

4. (4) follows directly from (1)-(3).

## 2.4 The definitions of symmetric normalized and random walk hypergraph Laplacians

The symmetric normalized hypergraph Laplacian (defined in [7,8,9]) is defined as follows
$$L_{sym} = I - D_v^{-\frac{1}{2}} HWD_e^{-1}H^T D_v^{-\frac{1}{2}}$$
The random walk hypergraph Laplacian (defined in [7,8,9]) is defined as follows
$$L_{rw} = I - D_v^{-1} HWD_e^{-1}H^T$$

## 2.5 Properties of $L_{sym}$ and $L_{rw}$

1. For every vector $f \in R^{|V|}$, we have
$$f^T L_{sym} f = \frac{1}{2}\sum_{e \in E}\sum_{\{u,v\} \subseteq E} \frac{w(e)}{d(e)}\left(\frac{f(u)}{\sqrt{d(u)}} - \frac{f(v)}{\sqrt{d(v)}}\right)^2$$

2. $\lambda$ is an eigenvalue of $L_{rw}$ with eigenvector u if and only if $\lambda$ is an eigenvalue of $L_{sym}$ with eigenvector $w = D_v^{\frac{1}{2}}u$

3. $\lambda$ is an eigenvalue of $L_{rw}$ with eigenvector u if and only if $\lambda$ and u solve the generalized eigen-problem $Lu = \lambda D_v u$

4. 0 is an eigenvalue of $L_{rw}$ with the constant one vector 1 as eigenvector. 0 is an eigenvalue of $L_{sym}$ with eigenvector $D_v^{\frac{1}{2}}1$

5. $L_{sym}$ is symmetric and positive semi-definite and $L_{sym}$ and $L_{rw}$ have $|V|$ non-negative real-valued eigenvalues $0 \leq \lambda_1 \leq \cdots \leq \lambda_{|V|}$

Proof:
1. The complete proof of (1) can be found in [7].
2. (2) can be seen easily by solving

$$L_{sym}w = \lambda w \Leftrightarrow \left(I - D_v^{-\frac{1}{2}} HWD_e^{-1}H^T D_v^{-\frac{1}{2}}\right)w = \lambda w$$
$$\Leftrightarrow D_v^{-\frac{1}{2}}\left(I - D_v^{-\frac{1}{2}} HWD_e^{-1}H^T D_v^{-\frac{1}{2}}\right)w = \lambda D_v^{-\frac{1}{2}}w$$
$$\Leftrightarrow D_v^{-\frac{1}{2}}w - D_v^{-1} HWD_e^{-1}H^T D_v^{-\frac{1}{2}}w = \lambda D_v^{-\frac{1}{2}}w$$

Let $u = D_v^{-\frac{1}{2}}w$, (in the other words, $w = D_v^{\frac{1}{2}}u$), we have

$$L_{sym}w = \lambda w \Leftrightarrow u - D_v^{-1}HWD_e^{-1}H^Tu = \lambda u$$
$$\Leftrightarrow (I - D_v^{-1}HWD_e^{-1}H^T)u = \lambda u$$
$$\Leftrightarrow L_{rw}u = \lambda u$$

This completes the proof.

3. (3) can be seen easily by solving

$$L_{rw}u = \lambda u \Leftrightarrow (I - D_v^{-1}HWD_e^{-1}H^T)u = \lambda u$$
$$\Leftrightarrow D_v(I - D_v^{-1}HWD_e^{-1}H^T)u = \lambda D_v u$$
$$\Leftrightarrow (D_v - HWD_e^{-1}H^T)u = \lambda D_v u$$
$$\Leftrightarrow Lu = \lambda D_v u$$

This completes the proof.

4. First, we need to prove that $L_{rw}1 = 0$.

Let $d_v \in R^{|V|}$ be the vector containing the degrees of vertices of hypergraph G, $d_e \in R^{|E|}$ be the vector containing the degrees of hyper-edges of hypergraph G, $w \in R^{|E|}$ be the vector containing the weights of hyper-edges of G, $1 \in R^{|V|}$ be vector of all ones, and $one \in R^{|E|}$ be the vector of all ones. Hence we have

$$L_{rw}1 = (I - D_v^{-1}HWD_e^{-1}H^T)1$$
$$= 1 - D_v^{-1}HWD_e^{-1}d_e$$
$$= 1 - D_v^{-1}HW one$$
$$= 1 - D_v^{-1}Hw$$
$$= 1 - D_v^{-1}d_v$$
$$= 0$$

The second statement is a direct consequence of (2).

5. The statement about $L_{sym}$ is a direct consequence of (1), then the statement about $L_{rw}$ is a direct consequence of (2).

## 3. ALGORITHMS

Given a set of a set of feature vectors of speech samples $\{x_1, \dots, x_l, x_{l+1}, \dots, x_{l+u}\}$ where $n = l + u$ is the total number of speech samples (i.e. vertices) in the hypergraph $G=(V,E)$ and given the incidence matrix $H$ of $G$. The method constructing $H$ from the data representing the set of speech samples will be described clearly in the Experiments and Results section.

Define $c$ be the total number of words and the matrix $F \in R^{n*c}$ be the estimated label matrix for the set of feature vectors of speech samples $\{x_1, \dots, x_l, x_{l+1}, \dots, x_{l+u}\}$, where the point $x_i$ is labeled as sign($F_{ij}$) for each word $j$ ($1 \leq j \leq c$). Please note that $\{x_1, \dots, x_l\}$ is the set of all labeled points and $\{x_{l+1}, \dots, x_{l+u}\}$ is the set of all un-labeled points.

Let $Y \in R^{n*c}$ the initial label matrix for $n$ speech samples in the hypergraph G be defined as follows

$$Y_{ij} = \begin{cases} 1 \text{ if } x_i \text{ belongs to word } j \text{ and } 1 \leq i \leq l \\ -1 \text{ if } x_i \text{ does not belong to word } j \text{ and } 1 \leq i \leq l \\ 0 \text{ if } l+1 \leq i \leq n \end{cases}$$

Our objective is to predict the labels of the un-labeled points $x_{l+1}, \ldots, x_{l+u}$. Basically, all speech samples in the same hyper-edge should have the same label.

### Random walk hypergraph Laplacian based semi-supervised learning algorithm

In this section, we will give the brief overview of the random walk hypergraph Laplacian based semi-supervised learning algorithm. The outline of this algorithm is as follows

1. Construct $D_v$ and $D_e$ from the incidence matrix H of G
2. Construct $S_{rw} = D_v^{-1}HWD_e^{-1}H^T$
3. Iterate until convergence
$F^{(t+1)} = \alpha S_{rw} F^{(t)} + (1-\alpha)Y$, where α is an arbitrary parameter belongs to [0,1]
4. Let $F^*$ be the limit of the sequence $\{F^{(t)}\}$. For each word j, label each speech sample $x_i$ ($l+1 \leq i \leq l+u$) as sign($F^*_{ij}$)

Next, we look for the closed-form solution of the random walk graph Laplacian based semi-supervised learning. In the other words, we need to show that

$$F^* = \lim_{t \to \infty} F^{(t)} = (1-\alpha)(I - \alpha S_{rw})^{-1}Y$$

Suppose $F^{(0)} = Y$. Thus, by induction,

$$F^{(t)} = \alpha^t S_{rw}^t Y + (1-\alpha)\sum_{i=0}^{t-1}(\alpha S_{rw})^i Y$$

Since $S_{rw}$ is the stochastic matrix, its eigenvalues are in [-1,1]. Moreover, since 0<α<1, thus

$$\lim_{t \to \infty} \alpha^t S_{rw}^t = 0$$

$$\lim_{t \to \infty} \sum_{i=0}^{t-1}(\alpha S_{rw})^i = (I - \alpha S_{rw})^{-1}$$

Therefore,

$$F^* = \lim_{t \to \infty} F^{(t)} = (1-\alpha)(I - \alpha S_{rw})^{-1}Y$$

Now, from the above formula, we can compute $F^*$ directly.

### Symmetric normalized hypergraph Laplacian based semi-supervised learning algorithm

Next, we will give the brief overview of the symmetric normalized hypergraph Laplacian based semi-supervised learning algorithm which can be obtained from [7,8,9]. The outline of this algorithm is as follows

1. Construct $D_v$ and $D_e$ from the incidence matrix H of G
2. Construct $S_{sym} = D_v^{-\frac{1}{2}} HWD_e^{-1}H^T D_v^{-\frac{1}{2}}$
3. Iterate until convergence
$F^{(t+1)} = \alpha S_{sym} F^{(t)} + (1-\alpha)Y$, where α is an arbitrary parameter belongs to [0,1]
4. Let $F^*$ be the limit of the sequence $\{F^{(t)}\}$. For each word j, label each speech sample $x_i$ ($l+1 \leq i \leq l+u$) as sign($F^*_{ij}$)

Next, we look for the closed-form solution of the normalized graph Laplacian based semi-supervised learning. In the other words, we need to show that

$$F^* = \lim_{t \to \infty} F^{(t)} = (1-\alpha)(I - \alpha S_{sym})^{-1}Y$$

Suppose $F^{(0)} = Y$. Thus, by induction

$$F^{(t)} = \alpha^t S_{sym}^t Y + (1-\alpha) \sum_{i=0}^{t-1} (\alpha S_{sym})^i Y$$

Since $S_{sym}$ is similar to $S_{rw}$ ($S_{rw} = D_v^{-1} HWD_e^{-1} H^T = D_v^{-\frac{1}{2}} S_{sym} D_v^{\frac{1}{2}}$) which is a stochastic matrix, eigenvalues of $S_{sym}$ belong to [-1,1]. Moreover, since 0<α<1, thus

$$\lim_{t \to \infty} \alpha^t S_{sym}^t = 0$$

$$\lim_{t \to \infty} \sum_{i=0}^{t-1} (\alpha S_{sym})^i = (I - \alpha S_{sym})^{-1}$$

Therefore,

$$F^* = \lim_{t \to \infty} F^{(t)} = (1-\alpha)(I - \alpha S_{sym})^{-1} Y$$

Now, from the above formula, we can compute $F^*$ directly.

**Un-normalized hypergraph Laplacian based semi-supervised learning algorithm**

Finally, we will give the brief overview of the un-normalized hypergraph Laplacian based semi-supervised learning algorithm. The outline of this algorithm is as follows

1. Construct $D_v$ and $D_e$ from the incidence matrix H of G
2. Construct $L = D_v - HWD_e^{-1} H^T$
3. Compute closed form solution $F^* = \gamma(L + \gamma I)^{-1} Y$, where $\gamma$ is any positive parameter
4. For each word j, label each speech samples $x_i$ ($l+1 \leq i \leq l+u$) as sign($F_{ij}^*$)

The closed form solution $F^*$ of un-normalized hypergraph Laplacian based semi-supervised learning algorithm will be derived clearly and completely in Regularization Framework section.

## 4. REGULARIZATION FRAMEWORKS

In this section, we will develop the regularization framework for the symmetric normalized hypergraph Laplacian based semi-supervised learning iterative version. First, let's consider the error function

$$E(F) = \frac{1}{2} \left\{ \sum_{e \in E} \sum_{\{u,v\} \subseteq E} \frac{w(e)}{d(e)} || \frac{F_u}{\sqrt{d(u)}} - \frac{F_v}{\sqrt{d(v)}} ||^2 \right\} + \gamma \sum_{i=1}^{|V|} ||F_i - Y_i||^2$$

In this error function $E(F)$, $F_i$ and $Y_i$ belong to $R^c$. Please note that c is the total number of words and $\gamma$ is the positive regularization parameters. Hence

$$F = \begin{bmatrix} F_1^T \\ \vdots \\ F_{|V|}^T \end{bmatrix} \text{ and } Y = \begin{bmatrix} Y_1^T \\ \vdots \\ Y_{|V|}^T \end{bmatrix}$$

Here $E(F)$ stands for the sum of the square loss between the estimated label matrix and the initial label matrix and the sum of the changes of a function F over the hyper-edges of the hypergraph [7,8,9].

Hence we can rewrite $E(F)$ as follows

$$E(F) = trace(F^T L_{sym} F) + \gamma trace((F-Y)^T (F-Y))$$

Our objective is to minimize this error function. In the other words, we solve

$$\frac{\partial E}{\partial F} = 0$$

This will lead to

$$\left(I - D_v^{-\frac{1}{2}}HWD_e^{-1}H^TD_v^{-\frac{1}{2}}\right)F + \gamma(F - Y) = 0$$

$$F - D_v^{-\frac{1}{2}}HWD_e^{-1}H^TD_v^{-\frac{1}{2}}F + \gamma F = \gamma Y$$

$$F - \frac{1}{1+\gamma}D_v^{-\frac{1}{2}}HWD_e^{-1}H^TD_v^{-\frac{1}{2}}F = \frac{\gamma}{1+\gamma}Y$$

$$\left(I - \frac{1}{1+\gamma}D_v^{-\frac{1}{2}}HWD_e^{-1}H^TD_v^{-\frac{1}{2}}\right)F = \frac{\gamma}{1+\gamma}Y$$

Let $\alpha = \frac{1}{1+\gamma}$. Hence the solution $F^*$ of the above equations is

$$F^* = (1-\alpha)(I - \alpha D_v^{-\frac{1}{2}}HWD_e^{-1}H^TD_v^{-\frac{1}{2}})^{-1}Y$$

Please note that $S_{rw} = D_v^{-1}HWD_e^{-1}H^T$ is not the symmetric matrix, thus we cannot develop the regularization framework for the random walk hypergraph Laplacian based semi-supervised learning iterative version.

Next, we will develop the regularization framework for the un-normalized hypergraph Laplacian based semi-supervised learning algorithms. First, let's consider the error function

$$E(F) = \frac{1}{2}\left\{\sum_{e \in E}\sum_{\{u,v\}\subseteq E}\frac{w(e)}{d(e)}||F_u - F_v||^2\right\} + \gamma \sum_{i=1}^{|V|}||F_i - Y_i||^2$$

In this error function $E(F)$, $F_i$ and $Y_i$ belong to $R^c$. Please note that $c$ is the total number of words and $\gamma$ is the positive regularization parameters. Hence

$$F = \begin{bmatrix} F_1^T \\ \vdots \\ F_{|V|}^T \end{bmatrix} \text{ and } Y = \begin{bmatrix} Y_1^T \\ \vdots \\ Y_{|V|}^T \end{bmatrix}$$

Here $E(F)$ stands for the sum of the square loss between the estimated label matrix and the initial label matrix and the sum of the changes of a function F over the hyper-edges of the hypergraph [7,8,9].

Hence we can rewrite $E(F)$ as follows

$$E(F) = F^TLF + \gamma\, trace((F-Y)^T(F-Y))$$

Please note that un-normalized hypergraph Laplacian matrix is $L = D_v - HWD_e^{-1}H^T$. Our objective is to minimize this error function. In the other words, we solve

$$\frac{\partial E}{\partial F} = 0$$

This will lead to

$$LF + \gamma(F - Y) = 0$$
$$(L + \gamma I)F = \gamma Y$$

Hence the solution $F^*$ of the above equations is

$$F^* = \gamma(L + \gamma I)^{-1}Y$$

Similarly, we can also obtain the other form of solution $F^*$ of the normalized graph Laplacian based semi-supervised learning algorithm as follows (note the symmetric normalized hypergraph Laplacian matrix is $L_{sym} = I - D_v^{-\frac{1}{2}}HWD_e^{-1}H^TD_v^{-\frac{1}{2}}$)

$$F^* = \gamma(L_{sym} + \gamma I)^{-1}Y$$

## 5. EXPERIMENTS AND RESULTS

In this paper, the set of 4,500 speech samples recorded of 50 different words (90 speech samples per word) are used for training. Then another set of 500 speech samples of these words are used for testing the sensitivity measure. This dataset is available from the IC Design lab at Faculty of Electricals-Electronics Engineering, University of Technology, Ho Chi Minh City. After being extracted from the conventional MFCC feature extraction method, the column sum of the MFCC feature matrix of the speech sample will be computed. The result of the column sum which is the $R^{26*1}$ column vector will be used as the feature vector of the three hyper-graph Laplacian based semi-supervised learning algorithms.

Normally, clustering methods offer a natural way to the problem identifying groups of feature vectors of speech samples that show very similar patterns and tend to have similar labels [7,8,9] in the given data. In this experiment, we use k-mean clustering method (i.e. the most popular "hard" clustering method). Then each cluster can be considered as the hyper-edge of the hypergraph. By using these hyper-edges, we can construct the incidence matrix $H$ of the hypergraph. To make things simple, we can set the number of clusters to be 250. In the other words, we must choose the number of clusters in order to maximize the sensitivity performance measures of the three hypergraph Laplacian based semi-supervised learning methods.

In this section, we experiment with the above seven methods in terms of sensitivity measure. All experiments were implemented in Matlab 6.5 on virtual machine. The sensitivity measure Q is given as follows

$$Q = \frac{True\ Positive}{True\ Positive + False\ Negative}$$

True Positive (TP), True Negative (TN), False Positive (FP), and False Negative (FN) are defined in the following table 1

**Table 1:** Definitions of TP, TN, FP, and FN

|  |  | Predicted Label | |
| --- | --- | --- | --- |
|  |  | Positive | Negative |
| Known Label | Positive | True Positive (TP) | False Negative (FN) |
|  | Negative | False Positive (FP) | True Negative (TN) |

In these experiments, parameter $\alpha$ is set to be 0.96 and $\gamma = 1$. For this dataset, the table 2 shows the sensitivity measures of the three graph based semi-supervised learning methods, HMM method (i.e. the current state of the art method of speech recognition application), and the three hypergraph based semi-supervised learning methods.

**Table 2:** Comparisons of symmetric normalized, random walk, and un-normalized graph Laplacian based methods, the symmetric normalized, random walk, and un-normalized hypergraph Laplacian based methods and HMM method

| Sensitivity Measures (%) | |
| --- | --- |
| HMM | 89 |
| Graph un-normalized | 97.60 |
| Graph random walk | 97.60 |
| Graph normalized | 97.60 |
| Hypergraph un-normalized | 98.40 |
| Hypergraph random walk | **98.60** |
| Hypergraph normalized | **98.60** |

The following figure 1 shows the sensitivity measures of the conventional HMM method and the three graph Laplacian based semi-supervised learning methods, and the three hypergraph Laplacian based semi-supervised learning methods:

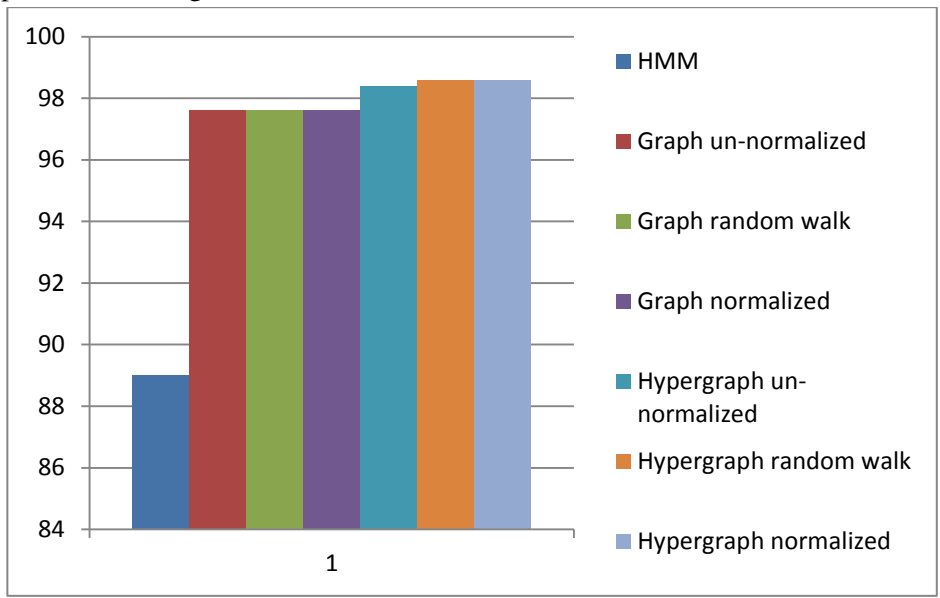

*Figure 1*. Sensitivity measures of the three graph based semi-supervised learning methods, three hyper-graph based semi-supervised learning methods and conventional HMM method

From the above table 2 and figure 1, we recognized that the symmetric normalized and random walk hyper-graph Laplacian based semi-supervised learning methods slightly perform better than the un-normalized hyper-graph Laplacian based semi-supervised learning method in terms of sensitivity measures. Moreover, these three hyper-graph Laplacian based semi-supervised learning methods outperform the current state of the art HMM method and the three graph based semi-supervised learning methods in speech recognition problem. First, since the graph based semi-supervised learning methods utilize the "pairwise relationship" between speech samples in the datasets (i.e. the kernel's definition) to build the predictive model, they must outperform the HMM method. Second, not only utilizing the "pairwise relationships" between speech samples, the three hypergraph Laplacian bases semi-supervised learning methods also utilizing the "complex relationships" among all speech samples; hence the three hypergraph based semi-supervised learning methods must outperform the three graph based semi-supervised learning methods.

## 6. CONCLUSIONS

We have proposed the detailed algorithms and regularization frameworks of the three un-normalized, symmetric normalized, and random walk hypergraph Laplacian based semi-supervised learning methods applying to speech recognition problem. Experiments show that these three methods outperform the graph based semi-supervised learning methods and the current state of the art HMM method since these three methods utilize the complex relationships among speech samples (i.e. not pairwise relationship). Moreover, these three methods can not only be used in the classification problem but also the ranking problem. In specific, given a set of genes (i.e. the queries) involved in a specific disease such as leukemia which is our future research, these three methods can be used to find more genes involved in leukemia by ranking genes in the hypergraph constructed from gene expression data. The genes with the highest rank can then be selected and checked by biology experts to see if the extended genes are in fact involved in leukemia. Finally, these selected genes will be used in cancer classification.

Recently, to the best of our knowledge, the un-normalized graph p-Laplacian based semi-supervised learning method have not yet been developed and applied to speech recognition problem. This method is worth investigated because of its difficult nature and its close connection to partial differential equation on graph field.

*Acknowledgement.* This work is funded by the Ministry of Science and Technology, State-level key program, Research for application and development of information technology and communications, code KC.01.23/11-15.